\documentclass[letterpaper, 10 pt, conference]{ieeeconf}

\IEEEoverridecommandlockouts
\usepackage{times}

\usepackage{booktabs}

\usepackage{multicol}
\usepackage[hidelinks,bookmarks=false]{hyperref}
\usepackage{amsmath} 
\usepackage{amssymb}
\usepackage{graphicx}
\usepackage[table]{xcolor}
\definecolor{oraclemethod}{RGB}{220,235,248}
\definecolor{probmethod}{RGB}{253,239,211}
\definecolor{abstractmethod}{RGB}{224,242,228}
\DeclareRobustCommand{\oracletypelabel}{\colorbox{oraclemethod}{\strut Oracle}}
\DeclareRobustCommand{\probtypelabel}{\colorbox{probmethod}{\strut Probabilistic}}
\DeclareRobustCommand{\abstracttypelabel}{\colorbox{abstractmethod}{\strut Abstraction}}
\usepackage{subcaption}

\usepackage{algorithm}
\usepackage{algorithmic}

\usepackage[acronym]{glossaries}

\usepackage{todonotes}
\usepackage{tikz}
\usetikzlibrary{arrows.meta,positioning,fit,calc}

\usepackage{multirow}
\usepackage{array}

\usepackage{caption}
\let\labelindent\relax
\usepackage{enumitem}

\usepackage[most]{tcolorbox}
\usepackage{cleveref}

\newcommand{\rev}[1]{\textcolor{black}{#1}}

\newlist{enumlite}{enumerate}{3}
\setlist[enumlite]{label=\arabic*),wide,labelindent=0pt}

\newlist{enumline}{enumerate*}{3}
\setlist[enumline]{label=(\arabic*), itemjoin*={{, and }}}

\newlist{itemlite}{itemize}{3}
\setlist[itemlite]{label=\textbullet,wide,labelindent=0pt,leftmargin=*}

\newacronym{HHZ}{HOPHY}{Hierarchical Off-Road Planning using
Hypergraphs}
\newacronym{mrta}{MRTA}{Multi-Robot Task Allocation}
\newglossaryentry{gsn}{name={GSNode}, plural={GSNodes}, description={Geometric--Semantic Node}}
\newglossaryentry{cr}{name={Coarse Region}, plural={Coarse Regions}, description={Coarse Region}}

\newglossaryentry{fr}{name={CLEAR},description={CLEAR}}

\glsdisablehyper

\makeglossaries

\newtcolorbox[auto counter, number within=section]{propertybox}[2][]{%
  colback=gray!5,
  colframe=black!60,
  boxrule=0.8pt,
  arc=3pt,
  left=6pt,
  right=6pt,
  top=6pt,
  bottom=6pt,
  fontupper=\scriptsize,
  title=Property~\thetcbcounter:~#2,
  label=#1
}

\hypersetup{
   pdfauthor={Author Names Omitted for Anonymous Review},
   pdftitle={HOPHY: A Hypergraph-Based Hierarchical Terrain Representation for Off-Road Path and Mission Planning},
   pdfsubject={Off-road path planning},
   pdfkeywords={Off-road planning, Hypergraphs, Learned cost model, Replanning}
}

\begin{document}

\title{\LARGE \bf HOPHY: A Hierarchical Hypergraph Representation for Off-Road Path and Mission Planning
}


\author{Pranay Meshram$^{1}$, Charuvahan Adhivarahan$^{1}$, 
Prithvi Poddar$^{2}$,
Ehsan Tarkesh Esfahani$^{2}$, Chen Wang$^{1}$,\\
 Souma Chowdhury$^{2}$, and Karthik Dantu$^{1}$
\thanks{$^{1}$Department of Computer Science and Engineering, University at Buffalo, NY 14260, USA.
{\tt\small \{pranaywa, charuvah, cwx, kdantu\}@buffalo.edu}}%
\thanks{$^{2}$Department of Mechanical and Aerospace Engineering, University at Buffalo, NY 14260, USA.
{\tt\small \{prithvid, ehsanesf, soumacho\}@buffalo.edu}}%
}



%

\maketitle
\thispagestyle{empty}
\pagestyle{empty}


\begin{abstract}
Mission-level autonomy for disaster response, search and rescue, and tactical UGV operations requires repeated path and mission planning as terrain conditions, agent types, and objectives change. Pixel-grid search is costly for repeated kilometer-scale queries, while semantic abstractions must maintain valid costs and connectivity as conditions change. We present HOPHY (Hierarchical Off-Road Planning using Hypergraphs), a reusable hierarchical terrain representation that organizes map-scale terrain into Geometrically connected Semantic regions (\glspl{gsn}), connectivity-preserving \glspl{cr}, and typed hyperedges for terrain, agent, and weather context. Hyperedge intersections select affected regions and incident edges for state updates without rebuilding the hierarchy. Across real off-road maps spanning kilometer-scale areas, HOPHY achieves 100\% planning success and $<0.01$\% median cost deviation from the oracle (pixel A*), with substantially lower query and replanning latency than the evaluated pixel and abstraction baselines. Applied to a multi-robot task-allocation (MRTA) problem, these gains reduce total computation by $79\times$ over pixel A* and $7.2\times$ over the fastest abstraction baseline, with mission makespan comparable to pixel A*. Finally, we demonstrate HOPHY on a physical Clearpath Jackal that successfully executes a 1.5-km, eight-task mission across mixed-surface outdoor terrain and a blockage-triggered replanned route.
\end{abstract}

\section{Introduction}
\label{sec:introduction}

Military logistics and search-and-rescue missions require ground robots to reach dispersed locations across large, unstructured landscapes, often where roads are unavailable or access has been disrupted~\cite{soares_argus_2025,murphy_disaster_2014}. Delivering supplies across mountainous terrain or reaching search sites after a disaster requires routes that account for vegetation, surface conditions, and slope, as well as the capabilities of each vehicle. Coordinating a team adds another demand: deciding which robot should visit each location, and in what order, requires repeatedly estimating the cost of travel between candidate tasks~\cite{doi:10.1177/0278364904045564,11163864}. These estimates must remain useful as rain changes traction, a route becomes blocked, or task assignments change. Such missions therefore need planning that combines terrain-aware route quality with fast, repeated queries over kilometer-scale areas and efficient updates as operating conditions evolve.

Much of the information needed to support this planning is available before deployment. Public landcover products, such as the North American Land Change Monitoring System (NALCMS)~\cite{nalcms2020}, identify surface classes, while digital elevation models, such as NASADEM~\cite{nasadem2020}, provide terrain geometry from which slope can be derived. Public weather observations and forecasts can supply additional context about precipitation and the areas it affects. Agent models complement these environmental layers by specifying mobility limits and relating terrain and operating conditions to traversal cost, travel time, or energy use. For example, the same rain-affected slope may remain traversable for one vehicle while becoming costly or infeasible for another. Translating these inputs into planning decisions requires a representation that preserves terrain connectivity and associates environmental changes with the regions and agents to which they apply.

\begin{figure}[t]
  \centering
  \includegraphics[width=\linewidth]{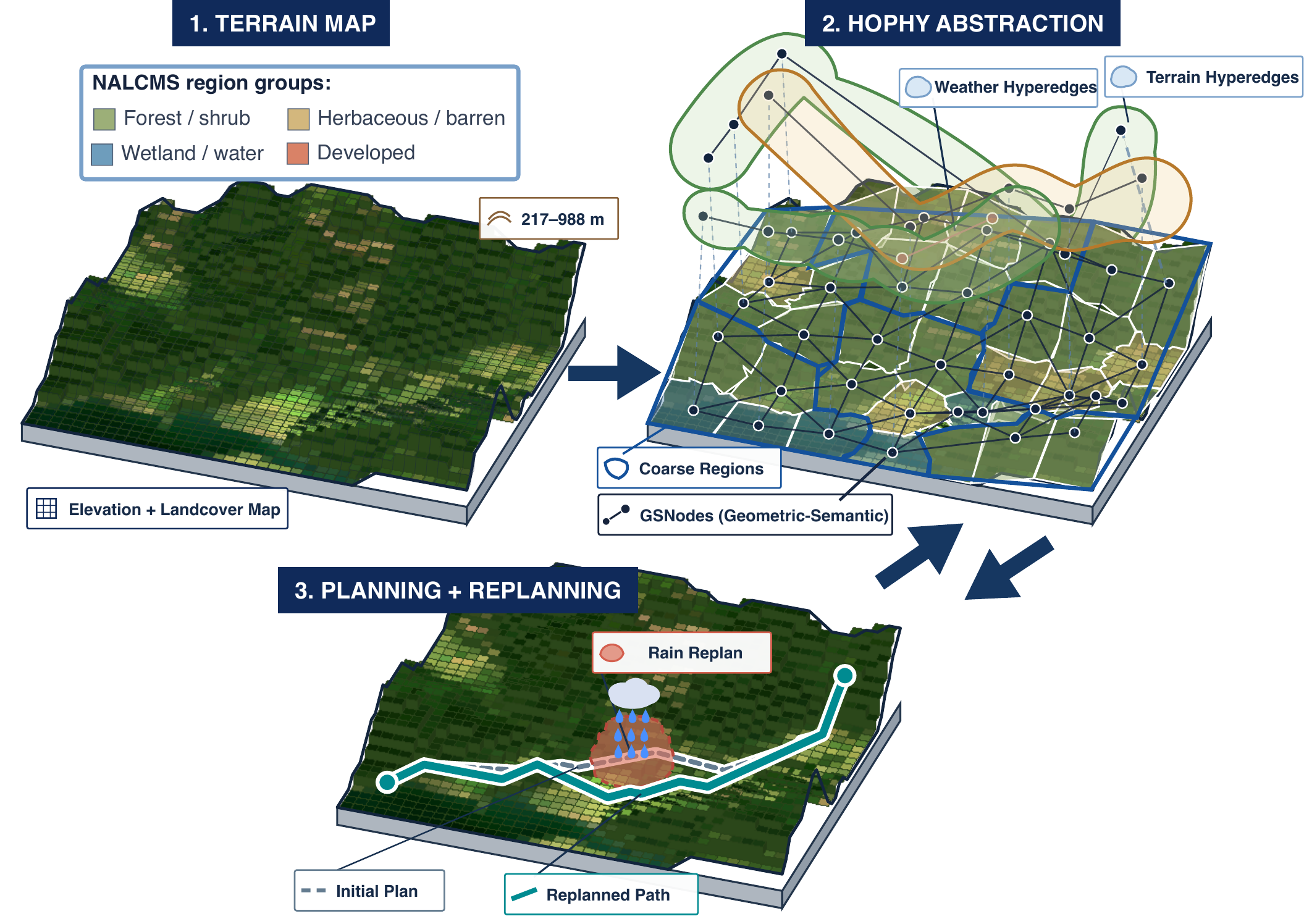}
  \caption{HOPHY on a Mount Rainier terrain subregion. Elevation and landcover are decomposed into GSNodes, Coarse Regions, and typed hyperedges for terrain-aware planning and localized replanning. Elevation vertically exaggerated.}
  \vspace{-2px}
  \label{fig:hophy_terrain_pipeline}
\end{figure}

The challenge is to make this representation efficient across an entire mission. Pixel-level search retains terrain detail, but repeatedly searching a large raster and evaluating traversal costs can dominate task-allocation and replanning time. 
Further, restricting repeated long-range searches within the abstraction and efficiently identifying which planning states must change under overlapping terrain, weather, and agent constraints is another challenge. A hierarchy can reduce the search space, while an explicit association between context and planning state can support targeted updates.

We present \gls{HHZ}, a reusable terrain representation that combines these capabilities. HOPHY organizes landcover and elevation into Geometrically connected Semantic regions (\glspl{gsn}) and connectivity-preserving \glspl{cr}, supporting coarse-to-fine routing followed by pixel-level refinement. Typed hyperedges encode overlapping terrain, agent-feasibility, and weather memberships over the same regions. Their intersections select the regions and incident connections whose costs, feasibility, or attributes must change, while retaining the underlying hierarchy for subsequent queries. This coupling of hierarchical search and contextual updates enables repeated path-cost estimation within mission optimization. \Cref{fig:hophy_terrain_pipeline} illustrates the representation. Our contributions are:

\begin{itemize}
    \item \textbf{A reusable hierarchical terrain abstraction} for repeated off-road path queries, combining connected semantic regions, coarse routing, and pixel-level refinement. Across the evaluated maps, median query times are 28--630\,ms, with 100\% planning success and less than 0.01\% median cost deviation from pixel A*.
    \item \textbf{A typed hyperedge update mechanism} that maps combinations of terrain, agent, and weather context to affected regions and connections. It supports rain and blockage updates without rebuilding the hierarchy, with mean update-plus-query latency below 400\,ms across the evaluated replanning conditions.
    \item \textbf{Validation from path queries to mission execution} on real terrain maps spanning 9--100\,km\textsuperscript{2}, including multi-agent task allocation and a physical ground robot demonstration. In the five-agent, fifty-task experiment, HOPHY reduces planning computation by $79\times$ relative to pixel A* and $7.2\times$ relative to CLEAR~\cite{meshram2026clear}, with estimated mission makespan comparable to pixel A*. A Clearpath Jackal executes a 1.5-km, eight-task mission and a blockage-triggered replacement route.
\end{itemize}

\section{Related Work}
\label{sec:related_work}

\noindent\textbf{Terrain abstractions and hierarchical planning.}
Off-road planning representations progressively reduce the search space from pixel grids to structured abstractions. Fixed-cluster abstractions~\cite{botea2004near} accelerate binary-grid search, but changing off-road costs invalidate cached connections. Flat semantic--geometric graphs~\cite{meshram2026clear} built from landcover and elevation substantially accelerate path queries and provide the closest flat-abstraction baseline in our evaluation. Hierarchical approaches~\cite{zheng_two-stage_2024,lee_trg-planner_2025} further stratify search through coarse-to-fine routing and risk-aware traversal over unstructured terrain. Context changes involving agents, weather, or blockages still require identifying and updating affected costs and connections. HOPHY maintains typed memberships for this identification alongside its reusable hierarchy.

\noindent\textbf{Dynamic replanning and traversal costs.}
Two lines of work address cost changes after initial planning but in orthogonal ways. Incremental search methods~\cite{koenig_improved_2002,lim_lazy_2024} repair shortest-path state when changed edges are supplied. Learned cost models~\cite{meng_terrainnet_2023,fu2025anynav} estimate surface-specific traversal costs from sensor observations, producing signals that a planner can consume. Repair methods~\cite{koenig_improved_2002,lim_lazy_2024} take changed edges as input, while cost models~\cite{meng_terrainnet_2023,fu2025anynav} produce per-surface estimates but do not organize the planning space. The proposed abstraction connects these roles: hyperedge intersections identify the affected planning edges and expose their updates to downstream search.

\noindent\textbf{Hypergraphs for repeated mission-planning queries.}
Hypergraphs have encoded natural-terrain relationships~\cite{fayek_using_1996} and coupled constraints in multi-robot task and motion planning~\cite{motes_hypergraph-based_2023}, exploiting hyperedges as constraint containers that group entities sharing a relationship. Their set-intersection semantics make them equally suited as overlap indices: given overlapping terrain, agent, and weather memberships, their intersection directly identifies the affected planning subset without scanning every terrain node. HOPHY uses these indices to maintain contextual state across repeated planning calls. MRTA solvers~\cite{doi:10.1177/0278364904045564} and course-of-action generators~\cite{11163864} expose the downstream demand for this capability, as both require repeated path-cost evaluations whose latency determines whether mission-level optimization is tractable at planning time. The proposed representation makes this inner loop fast enough to serve those repeated evaluations.

\section{Hierarchical Representation with Hypergraphs}
\label{sec:method}
\glsreset{gsn}
\glsreset{cr}

\begin{figure*}[t]
  \centering
  \vspace{-12pt}
  \includegraphics[width=0.8\textwidth]{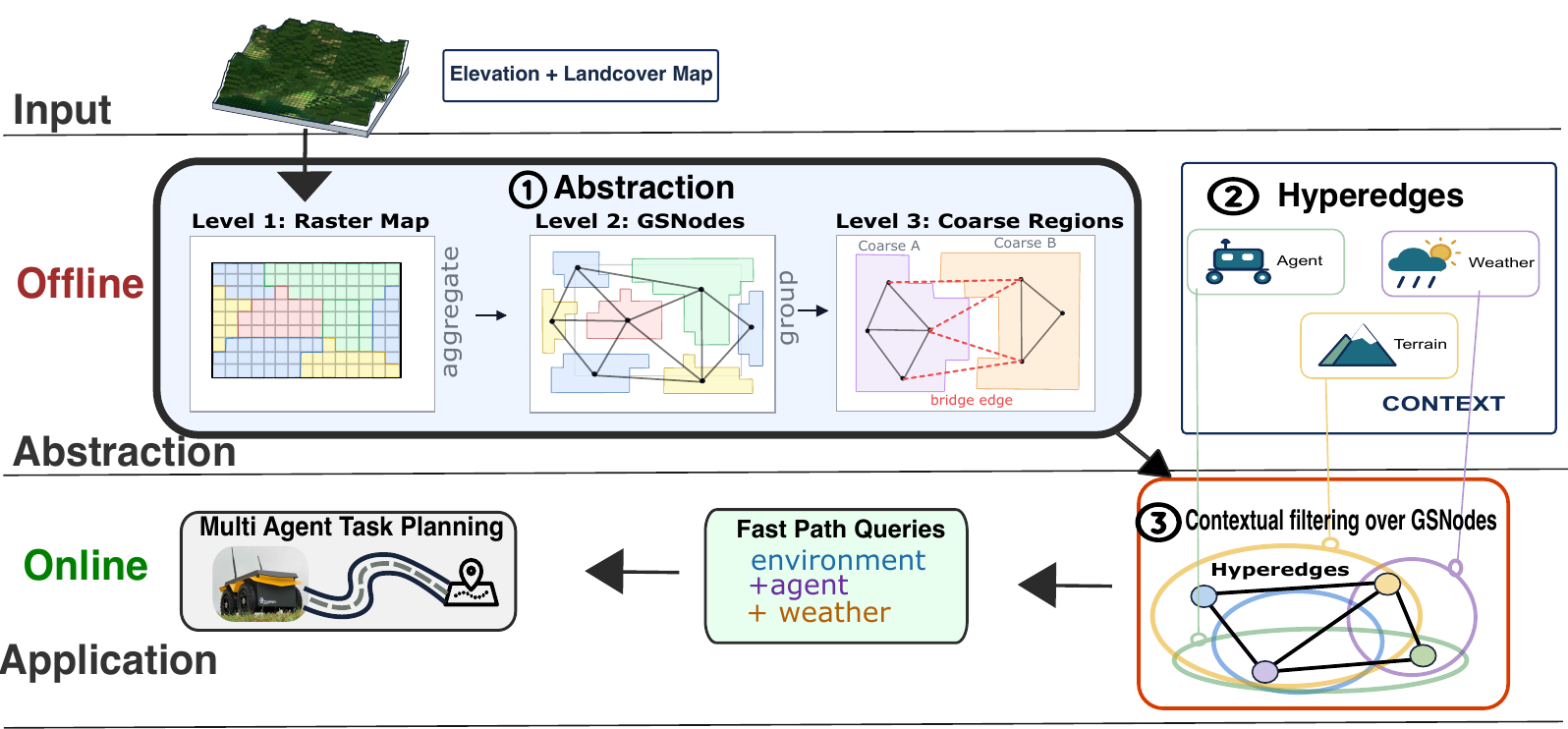}
  \vspace{-4pt}
  \caption{\rev{HOPHY terrain hierarchy and query pipeline. Terrain inputs are decomposed into \glspl{gsn} and \glspl{cr}. Typed hyperedges index overlapping contextual memberships for localized updates, repeated path queries, and multi-agent planning.}}
  \vspace{-16pt}
  \label{fig:HHZ_framework}
\end{figure*}

HOPHY is constructed offline in two stages, shown in \Cref{fig:HHZ_framework} and summarized in Algorithm~\ref{alg:build_hierarchy}. First, terrain maps are converted into a spatial hierarchy of GSNodes and Coarse Regions, with adjacency and bridge edges preserving connectivity. Second, typed hyperedges associate these GSNodes with terrain properties, agent feasibility constraints, and weather conditions. The following subsections describe these stages in order. The resulting representation supports the online path queries and contextual updates described in \Cref{sec:ma_planning}.

\subsection{Hierarchical Representation Building}
\label{subsec:env_rep}

\noindent\textbf{Terrain preprocessing:}
At Level~1 (Raster Map) in \Cref{fig:HHZ_framework}, cells are labeled by landcover and local grade to identify similar patches. \textsc{CombineTerrain} forms joint class labels from landcover classes and elevation-derived grade bins. Let $\mathcal{L}$ be the landcover classes and $\mathcal{B}_E$ the grade bins computed from elevation $E$. Each cell $x$ receives a label encoding the pair $(\ell(x),b_E(x))\in\mathcal{L}\times\mathcal{B}_E$. Cells sharing a joint label therefore have the same landcover class and approximately similar local grade. In our implementation, eight landcover classes are combined with three grade bins, $[0,10]\%$, $(10,20]\%$, and $(20,35]\%$, using
\[
M(x)=3\bigl(\ell(x)-1\bigr)+b_E(x)+1,
\]
where $\ell(x)\in\{1,\ldots,8\}$ and $b_E(x)\in\{0,1,2\}$. This produces 24 joint labels, with a separate label 25 for all terrain above $35\%$ grade. These labels guide segmentation by landcover and local terrain geometry, rather than by absolute elevation.

Morphological opening removes small isolated patches, and closing fills small gaps (\textsc{MorphSimplify}). Applying these operations at scale $\alpha_f$ produces $M_f$, whose connected patches define the fine-segment set $\mathcal{S}$ (\textsc{ConnectedSegments}). Fine segments serve only as intermediate construction units.

\noindent\textbf{Geometrically connected Semantic regions (\Glspl{gsn})}:
Repeated pixel-level search is expensive in unstructured terrain. A \gls{gsn} provides a larger planning unit by merging connected fine segments with the same joint landcover--grade label from $\mathcal{L}\times\mathcal{B}_E$. The Level~2 graphic in \Cref{fig:HHZ_framework} illustrates this grouping of fine segments into GSNodes. For a connected group $B\subseteq\mathcal{S}$ of fine segments sharing a joint label, the resulting node is
\[
D_B=\bigcup_{s\in B}D_s,\qquad g_B=(D_B,m_B,\bar z_B),
\]
where $D_s$ is a segment's geometry, $m_B$ is the shared joint label, and $\bar z_B$ is the area-weighted mean elevation. Thus, each GSNode is connected and has a common landcover class and grade bin, while its shape may be nonconvex. Let $\mathcal{G}$ denote the complete GSNode set.

\noindent\textbf{Coarse Regions}:
The GSNode graph can still contain tens of thousands of nodes. \Glspl{cr} restrict long-range search to a coarse corridor, with bridge edges preserving connectivity between neighboring regions (Level~3 in \Cref{fig:HHZ_framework}). 

Further simplification at scale $\alpha_c$ produces $M_c$, whose boundaries define coarse polygons $\mathcal{P}$ after merging small polygons using threshold $A_{\min}$ (\textsc{CoarsePolygons}). During construction, fine segments are assigned to each polygon $P\in\mathcal{P}$ by overlap or centroid containment, with each segment assigned once (\textsc{AssignedSegments}). Segments with matching joint labels are merged within that polygon (\textsc{MergeByTerrain}), producing $\mathcal{G}_P$. This ensures that each GSNode has a unique coarse parent. A nonempty polygon and its nodes define a \gls{cr} (\textsc{CreateCoarseRegion}) as
\[
c_P=(P,\mathcal{G}_P),\qquad
\mathcal{C}=\{c_P:P\in\mathcal{P},\ \mathcal{G}_P\neq\emptyset\}.
\]
The parent mapping $\pi_{G\rightarrow C}(g)=c_P$ for $g\in\mathcal{G}_P$ assigns each GSNode to exactly one Coarse Region, and every retained region is nonempty.

\noindent\textbf{Adjacency across abstractions:}
\Gls{gsn} adjacency defines the \textit{GSGraph}, represented by the directed graph $G_G=(\mathcal{G},E_G)$, where $(g_i,g_j)\in E_G$ if distinct \glspl{gsn} $g_i$ and $g_j$ share a boundary. Both directions are included, while their runtime costs may differ. Contraction under $\pi_{G\rightarrow C}$ yields $G_C=(\mathcal{C},E_C)$, where
\[
\begin{aligned}
(c_k,c_\ell)\in E_C
&\iff c_k\neq c_\ell \ \wedge\ \exists(g_i,g_j)\in E_G:\\
&\quad \pi_{G\rightarrow C}(g_i)=c_k \ \wedge\
\pi_{G\rightarrow C}(g_j)=c_\ell.
\end{aligned}
\]

\noindent\textbf{Offline construction.}
\label{subsec:hconst}
Algorithm~\ref{alg:build_hierarchy} implements these definitions in one offline pass. \textsc{BuildGSAdjacency} connects neighboring GSNodes to form $G_G$. \textsc{BuildBridgeGraph} links their parent coarse regions across boundary edges to form $G_C$. Finally, \textsc{IndexSets} groups GSNodes by terrain, agent feasibility, and weather context into the hyperedges defined next. The hierarchy is retained for online queries and updates.

\begin{algorithm}[t]
\caption{\rev{Build HOPHY Hierarchy}}
\label{alg:build_hierarchy}
\footnotesize
\begin{algorithmic}[1]
\setlength{\itemsep}{0pt}
\setlength{\parsep}{0pt}
\setlength{\parskip}{0pt}
\setlength{\baselineskip}{9pt}
\REQUIRE Maps $L,E$, scales $\alpha_f,\alpha_c$, area threshold $A_{\min}$, agent constraints $\mathcal{F}$, weather $q_0$
\ENSURE Graphs $G_G,G_C$ and typed hyperedges $\mathcal{E}_H$
\STATE Form joint terrain labels $M$ from $L$ and $E$.
\STATE Simplify $M$ at $\alpha_f$, then $\alpha_c$, yielding $M_f,M_c$.
\STATE Extract fine segments $\mathcal{S}$ from $M_f$ and coarse polygons $\mathcal{P}$ from $M_c$ using $A_{\min}$.
\STATE Assign each fine segment to one coarse polygon.
\STATE Within each polygon, merge connected segments with the same joint label into GSNodes $\mathcal{G}_P$.
\STATE Form $\mathcal{G}=\bigcup_P\mathcal{G}_P$ and nonempty Coarse Regions $\mathcal{C}=\{(P,\mathcal{G}_P):\mathcal{G}_P\neq\emptyset\}$.
\STATE Build GSNode adjacency $G_G$ and coarse bridge graph $G_C$.
\STATE Index terrain, agent ($\mathcal{F}$), and weather ($q_0$) memberships as $\mathcal{E}_H$.
\RETURN $G_G,G_C,\mathcal{E}_H$
\end{algorithmic}
\end{algorithm}

\subsection{Hyperedge Construction}
\label{subsec:hypergraph_def}
\label{subsec:hyperedge_construction}

The Hyperedges block in \Cref{fig:HHZ_framework} associates GSNodes with overlapping terrain, agent-feasibility, and weather memberships, without requiring spatial connectivity.

The typed hypergraph is $\mathcal{H}=(\mathcal{V},\mathcal{E}_H,\lambda,\mathbf{X})$, where $\mathcal{V}=\mathcal{G}$, $\mathcal{E}_H\subseteq 2^{\mathcal{V}}$ is the set of hyperedges, $\lambda:\mathcal{E}_H\rightarrow\mathcal{T}$ assigns each hyperedge a contextual type, and $\mathbf{X}:\mathcal{V}\rightarrow\mathcal{A}$ maps \glspl{gsn} to attributes. Membership $v\in e$ defines incidence. Hyperedge intersections select nodes for contextual updates, while $E_G$ retains spatial adjacency.

We construct three membership families over the same \glspl{gsn}:

\noindent\textbf{Terrain-based hyperedges} ($e_{\text{landcover}}$, $e_{\text{grade}}$) group GSNodes by landcover class or the grade intervals defined above.

\noindent\textbf{Agent-based hyperedges} ($e_{\text{agent}}$) collect GSNodes satisfying each agent's feasibility limits, such as the $35\%$ grade limit for wheeled vehicles. Built after hierarchy construction, these memberships are selected for the querying agent online. Traversal costs come from a separate function (\Cref{subsec:path_planning}).

\noindent\textbf{Weather-effect hyperedges} ($e_{\text{weather}}$) collect GSNodes exposed to conditions such as rain, identifying where context-dependent traversal costs change.

\noindent\textbf{Contextual updates through set operations.}
\rev{The Contextual filtering block in \Cref{fig:HHZ_framework} selects nodes for a typed update operator
$\Omega=(\tau,\{e_i\}_{i=1}^{k},\phi_V,\phi_E)$, where $\tau$ declares the update type (cost, feasibility, or attribute relabel), the $e_i\in\mathcal{E}_H$ are applicability memberships, and $\phi_V$ and $\phi_E$ are node- and edge-state transformations. Its support is}
\[
U(\Omega)=\bigcap_{i=1}^{k}e_i
\quad\text{e.g.,}\quad
e_{\text{agent}} \cap e_{\text{weather}} \cap e_{\text{landcover}} \cap e_{\text{grade}},
\]
\rev{and only \glspl{gsn} in $U(\Omega)$ and their incident directed edges are transformed. Operators on disjoint attributes commute. When operators target the same attribute, a declared precedence applies (feasibility masking before cost modification). Memberships are stored as hash sets. With $E(U)=\{(u,v)\in E_G:u\in U\vee v\in U\}$, the set-selection and state-mutation cost is $O(\sum_i|e_i|+|U|+|E(U)|)$ rather than a full $O(|\mathcal{G}|+|E_G|)$ scan. This becomes a full-graph operation when every \gls{gsn} is matched. Cache invalidation is described in \Cref{subsec:path_planning}.}

\section{Path and Mission Planning}
\label{sec:ma_planning}
The constructed hierarchy supports rapid online route planning. We first explain how coarse-to-fine path planning uses this hierarchy under agent and environmental constraints, as summarized in Algorithm~\ref{alg:coarse_pathfinding}. We then use this approach to estimate inter-task travel costs and solve a multi-robot task allocation (MRTA) problem for a heterogeneous team.

\subsection{Path Planning with \gls{HHZ}}
\label{subsec:path_planning}
Algorithm~\ref{alg:coarse_pathfinding} implements the Fast Path Queries block in \Cref{fig:HHZ_framework}. It maps endpoints to valid GSNodes and parent Coarse Regions, preferring representatives in the endpoints' raster-connected components. After contextual filtering, one connected GSNode search follows a coarse corridor and widens it if needed. Coarse routes and search subsets are cached. Directed pixel-level A* then connects the original endpoints within the seeded corridor under the current mask and online traversal costs.

\noindent\textbf{Partial blockages and cache invalidation.}
Blockages mask affected pixels and exclude GSNodes exceeding the configured blocked-pixel count and fraction thresholds. Endpoint GSNodes remain candidates, raster refinement enforces partial blockages and pixel connectivity. Rain updates change friction-dependent features and clear dynamic transition, abstract-edge, and compiled inference caches, retaining the hierarchy and static descriptors. Query-index refreshes clear query-result and GSNode-subproblem caches and, when requested, route caches. Invalidation is not limited to affected edges.

\begin{algorithm}[t]
\caption{\rev{Coarse-to-Fine HOPHY Query}}
\label{alg:coarse_pathfinding}
\footnotesize
\begin{algorithmic}[1]
\setlength{\itemsep}{0pt}
\setlength{\parsep}{0pt}
\setlength{\parskip}{0pt}
\setlength{\baselineskip}{9pt}
\REQUIRE Start $x_s$, goal $x_g$, graphs $G_G,G_C$, context $q$, cache $\mathcal{K}$
\ENSURE Feasible flag, path $\pi$, cost $J$
\STATE \textcolor{gray}{\textit{// Map query endpoints to the maintained hierarchy}}
\STATE $(g_s,c_s)\leftarrow\textsc{Locate\gls{gsn}}(x_s)$
\STATE $(g_g,c_g)\leftarrow\textsc{Locate\gls{gsn}}(x_g)$
\STATE \textcolor{teal}{\textit{// Apply weather, blockage, and agent constraints}}
\STATE $G_G^q\leftarrow\textsc{ApplyHyperedgeFilter}(G_G,q)$
\IF{$g_s=\emptyset$ or $g_g=\emptyset$}
  \RETURN \textbf{false}, $\emptyset$, $\infty$
\ENDIF
\STATE \textcolor{orange}{\textit{// Use local search when no coarse transition is needed}}
\IF{$c_s=c_g$}
  \STATE $\sigma\leftarrow\textsc{\gls{gsn}AStar}(G_G^q,g_s,g_g,q)$
\ELSE
  \STATE \textcolor{purple}{\textit{// Search a connected corridor across coarse boundaries}}
  \STATE $\rho\leftarrow\textsc{CoarseRoute}(G_C,c_s,c_g,\mathcal{K})$
  \STATE $\sigma\leftarrow\textsc{CorridorGSNodeSearch}(G_G^q,g_s,g_g,\rho,\mathcal{K})$
  \STATE \COMMENT{Widen to neighboring coarse regions if needed}
\ENDIF
\IF{$\sigma=\emptyset$}
  \RETURN \textbf{false}, $\emptyset$, $\infty$
\ENDIF
\STATE $\mathcal{C}_q\leftarrow\textsc{RasterCorridor}(\sigma,x_s,x_g,q)$
\STATE $(\pi,J)\leftarrow\textsc{DirectedAStar}(\mathcal{C}_q,x_s,x_g,q)$
\RETURN $(\pi\neq\emptyset)$, $\pi$, $J$
\end{algorithmic}
\end{algorithm}

\noindent\textbf{Traversal Cost Interface:}
\label{sec:cost_model}
A representation-agnostic function supplies directed weights $w(u,v\mid q)$ from transition descriptors and agent--environment context. Signed grade and terrain transitions generally make $w(u,v\mid q)\neq w(v,u\mid q)$, so one scalar cost raster cannot represent the objective. Full-grid A* and HOPHY evaluate the same function online, with HOPHY restricting the evaluation region.

\subsection{Mission Planning}
\label{sec:mrta}

For the Multi Task Planning block in \Cref{fig:HHZ_framework}, we use \gls{HHZ} inside a heterogeneous Min-Max mTSP formulation~\cite{francca1995m} following the ST-SR MRTA taxonomy~\cite{doi:10.1177/0278364904045564}. The objective is to minimize the latest agent completion time, $\min_j \max T_{a_j}$, over tasks in a large off-road environment. We solve the allocation with a genetic algorithm~\cite{11163864}. \gls{HHZ} supplies the repeated inter-task travel-time and energy estimates.

\section{Evaluation}
\label{sec:evaluation}
We evaluate the proposed representation across increasingly demanding settings: repeated long-range path queries, replanning after terrain changes, mechanism and scaling analysis, multi-agent mission planning, and physical Jackal execution. This progression is organized around five questions:

\begin{tcolorbox}[
  colback=gray!5,
  colframe=black!60,
  boxrule=0.8pt,
  arc=3pt,
  fontupper=\footnotesize,
  left=4pt,
  right=4pt,
  top=3pt,
  bottom=3pt
]

\phantomsection
\textbf{\rev{Q1: Path-planning performance}}:\label{prop:path_performance}
\rev{Does HOPHY accelerate repeated queries without sacrificing success or path quality?}

\phantomsection
\textbf{\rev{Q2: Replanning}}:\label{prop:replanning}
\rev{Does this advantage persist after terrain costs or feasibility change?}

\phantomsection
\textbf{\rev{Q3: Mechanisms}}:\label{prop:mechanisms}
\rev{Which components produce the gains, and how do hyperedge updates scale?}

\phantomsection
\textbf{\rev{Q4: Mission planning}}:\label{prop:mission_planning}
\rev{Do faster queries improve a multi-agent optimization loop?}

\phantomsection
\textbf{\rev{Q5: Deployment}}:\label{prop:deployment}
\rev{Can the resulting paths be executed on a physical robot?}

\end{tcolorbox}

\subsection{Experimental Protocol}
\label{sec:experimental_protocol}

\noindent\textbf{Platform and maps.}
All computational experiments use Ubuntu 20.04 on a 12th Gen Intel Core i9-12900K with 32\,GB RAM. We evaluate Wharton~(W, 9\,km\textsuperscript{2}, $355\!\times\!268$), Humphrey~(H, 50\,km\textsuperscript{2}, $1707\!\times\!1670$), and Rainier~(R, 100\,km\textsuperscript{2}, $3198\!\times\!2613$). Each map combines NASA DEM~\cite{nasadem2020} elevation and NALCMS~\cite{nalcms2020} landcover with a nominal source resolution of 30\,m. We generate 90 distance-stratified start--goal pairs per map (270 total).

\noindent\textbf{Hierarchy parameters.}
\Cref{tab:hierarchy_settings} lists the morphological ablation settings for Algorithm~\ref{alg:build_hierarchy}. Scales are in pixels and areas in pixel$^2$. A zero $\alpha_f$ disables fine smoothing. $A_{\mathrm{fine}}$ filters fine regions, and $A_{\min}$ controls coarse-polygon merging. The verified path-query results instead use a direct-raster hierarchy that bypasses fine-region construction.

\begin{table}[t]
\centering
\scriptsize
\setlength{\tabcolsep}{4pt}
\renewcommand{\arraystretch}{0.9}
\caption{Abstraction hyperparameters.}
\label{tab:hierarchy_settings}
\begin{tabular}{lrrrr}
\toprule
Map & $A_{\mathrm{fine}}$ & $\alpha_f$ & $\alpha_c$ & $A_{\min}$ \\
\midrule
Wharton & 0 & 0 & 4 & 100 \\
Humphrey & 4 & 2 & 16 & 500 \\
Rainier & 4 & 4 & 16 & 1000 \\
\bottomrule
\end{tabular}
\vspace{-7pt}
\end{table}

\noindent\textbf{Query and cost settings.}
Verified queries start in dry weather ($q_0$), with target speed $1.0\,\mathrm{m/s}$ and maximum grade $35\%$. Cost, risk, time, and step weights are 8, 4, 0.02, and 1. Pixel refinement uses one corridor of radius 192 pixels. Compilation precedes timing, with no graph-query warmups. The rain-replanning runner excludes GSNodes with at least 4 blocked pixels and blocked fraction 0.08, uses zero blockage-expansion hops, and retains endpoint GSNodes for raster refinement.


\noindent\textbf{Objective and baselines.}
All planners use identical terrain inputs and a shared learned Jackal traversal objective. Taking inspiration from TERP~\cite{weerakoon_terp_2022}, we train an MLP with two shared tanh layers (10 and 48 neurons) that predicts traversal cost from grade, roughness, friction, target speed, and distance, trained on 18{,}090 augmented Jackal traversal samples from BeamNG.tech, a soft-body vehicle simulation platform~\cite{maul_beamng_2021}. All methods use this same cost function to score transitions. Baselines are 2-stage PRM+A*~\cite{zheng_two-stage_2024}, CLEAR~\cite{meshram2026clear}, RRT*, and RRT-Connect.

\noindent\textbf{Reporting.}
Because a single scalar measure can obscure complementary failure modes in geometric evaluation~\cite{Meshram_2026_WACV}, we report success, cost deviation, length deviation, and query time separately rather than collapsing them into one score.
All methods were evaluated on the same hardware using identical start--goal pairs and a shared traversal-cost model. Query time includes required search-time model evaluations. RRT* and RRT-Connect use 3000 and 10000 iterations, respectively, with three attempts. Cost and length use the best feasible attempt, while time includes all attempts. Their low Rainier success (12/90 and 20/90) makes their deltas descriptive stress-test results, and they are excluded from replanning because every changed-cost query rebuilds the tree.

\subsection{Path-Planning Performance}
\label{sec:path_benchmark}

We first address Q1 by comparing success, query time, and path quality over the shared 270-query benchmark. As summarized in~\Cref{tab:comprehensive_efficiency}, the direct HOPHY implementation succeeds on every query and records median query times of 28, 250, and 630\,ms on Wharton, Humphrey, and Rainier, respectively. These correspond to speedups of $71.4\times$, $249.1\times$, and $146.6\times$ over oracle pixel A*, and $4.9\times$, $23.7\times$, and $12.3\times$ over CLEAR, the fastest abstraction baseline.

Our abstraction has zero median cost and length deviation on every map and small mean deviations on Rainier, indicating close agreement with pixel A* across the evaluated queries.

\begin{figure*}[t]
  \centering
  \includegraphics[width=\textwidth]{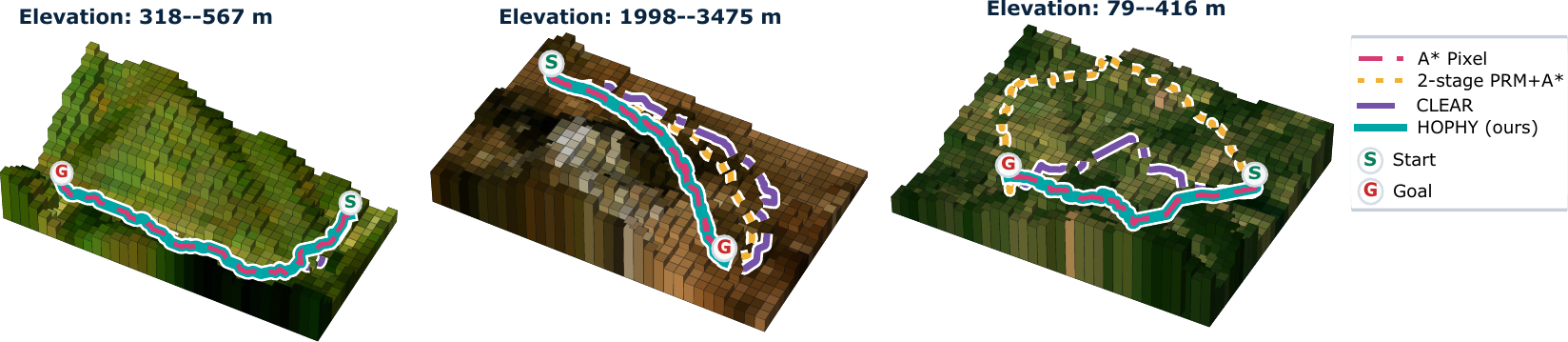}
  \caption{Representative routes on real stepped-elevation terrain (Wharton, Humphrey, Rainier). HOPHY follows the pixel-A* corridor. CLEAR and 2-stage PRM+A* select different terrain corridors. Elevation vertically exaggerated.}
  \label{fig:qualitative_paths}
\end{figure*}

Representative routes in~\Cref{fig:qualitative_paths} show this behavior directly across the three terrain scales.

\begin{figure}
\centering
\includegraphics[width=0.8\linewidth]{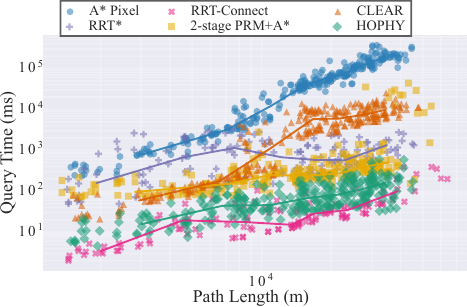}
\vspace{-6pt}
\caption{Query time vs.\ path length}
\label{fig:distance_time_efficiency}
\end{figure}

The distance-stratified view in~\Cref{fig:distance_time_efficiency} shows how query time changes with route length. Pixel A* scores every expanded transition, whereas HOPHY limits computation to abstract edges requested by search and transitions inside the selected refinement corridor. \Cref{tab:planner_initialization} reports the best verified direct-hierarchy build plus static-descriptor preparation time. The HOPHY state contains 11{,}993, 27{,}666, and 51{,}217 \glspl{gsn}, and 134, 962, and 1025 \glspl{cr}.

\begin{table}[t]
\centering
\scriptsize
\setlength{\tabcolsep}{4pt}
\renewcommand{\arraystretch}{0.9}
\caption{Initialization time (s). HOPHY initialization includes hierarchy construction and static-descriptor preparation, excluding compilation and path queries.}
\label{tab:planner_initialization}
\begin{tabular}{lrrr}
\toprule
\textbf{Method} & \textbf{W} & \textbf{H} & \textbf{R} \\
\midrule
\rev{CLEAR} & \rev{19.91} & \rev{67.36} & \rev{111.84} \\
\rev{2-stage PRM+A*} & \rev{4.86} & \rev{5.65} & \rev{8.34} \\
\rev{HOPHY} & \rev{\textbf{1.306}} & \rev{\textbf{5.389}} & \rev{20.030} \\
\midrule
\rev{Initialization / mean query time} & \rev{\textbf{28}} & \rev{\textbf{20}} & \rev{31} \\
\bottomrule
\end{tabular}
\vspace{-7pt}
\end{table}

\begin{table}[t]
\centering
\tiny
\setlength{\tabcolsep}{1.4pt}
\renewcommand{\arraystretch}{0.82}
\caption{Path-planning comparison over 90 start--goal pairs per map. Method types: \oracletypelabel, \probtypelabel, and \abstracttypelabel. Cost delta, length delta, and query time are mean$\pm$std/median relative to oracle (pixel A*).}
\label{tab:comprehensive_efficiency}
\resizebox{\columnwidth}{!}{%
\begingroup
\begin{tabular}{l c c c c}
\toprule
\textbf{Method} & \textbf{Succ.} & \textbf{$\Delta$Cost} & \textbf{$\Delta$Len.} & \textbf{Query} \\
 &  & (\%) & (\%) & (ms) \\
\midrule
\multicolumn{5}{l}{\textbf{Wharton (W)}} \\
\cellcolor{oraclemethod} A* & 90/90 & --- & --- & 2686.29$\pm$2394.74 / 1996.84 \\
\cellcolor{probmethod} RRT* & 60/90 & 38.82$\pm$17.92 / 35.32 & -2.59$\pm$3.16 / -2.39 & 3453.05$\pm$2593.50 / 3044.74 \\
\cellcolor{probmethod} RRT-C & 47/90 & 58.21$\pm$27.16 / 57.04 & 7.99$\pm$9.33 / 7.33 & 232.06$\pm$359.86 / 108.42 \\
\cellcolor{abstractmethod} \gls{fr} & 90/90 & 2.72$\pm$1.57 / 2.31 & -1.92$\pm$4.78 / 0.00 & 182.66$\pm$208.44 / 135.84 \\
\cellcolor{abstractmethod} 2-stage PRM+A* & 90/90 & \textbf{0.00$\pm$0.00 / 0.00} & \textbf{0.00$\pm$0.00 / 0.00} & 655.96$\pm$386.12 / 582.21 \\
\cellcolor{abstractmethod} HOPHY & 90/90 & \textbf{0.00$\pm$0.00 / 0.00} & \textbf{0.00$\pm$0.00 / 0.00} & \textbf{47.36$\pm$93.29 / 27.95} \\
\midrule
\multicolumn{5}{l}{\textbf{Humphrey (H)}} \\
\cellcolor{oraclemethod} A* & 90/90 & --- & --- & 87726.10$\pm$73544.29 / 62256.64 \\
\cellcolor{probmethod} RRT* & 72/90 & 33.77$\pm$18.45 / 29.91 & 10.03$\pm$16.49 / 4.62 & 3544.90$\pm$2561.50 / 2878.57 \\
\cellcolor{probmethod} RRT-C & 61/90 & 38.04$\pm$29.22 / 24.43 & 11.97$\pm$23.83 / 0.33 & 284.58$\pm$366.61 / 119.13 \\
\cellcolor{abstractmethod} \gls{fr} & 90/90 & 14.69$\pm$3.04 / 14.49 & 3.98$\pm$3.20 / 3.61 & 6502.94$\pm$4093.79 / 5918.23 \\
\cellcolor{abstractmethod} 2-stage PRM+A* & 90/90 & 4.14$\pm$6.21 / 1.20 & -3.02$\pm$5.41 / -0.73 & 9720.27$\pm$4201.28 / 9058.88 \\
\cellcolor{abstractmethod} HOPHY & 90/90 & \textbf{0.00$\pm$0.00 / 0.00} & \textbf{0.00$\pm$0.00 / 0.00} & \textbf{274.82$\pm$116.99 / 249.88} \\
\midrule
\multicolumn{5}{l}{\textbf{Rainier (R)}} \\
\cellcolor{oraclemethod} A* & 90/90 & --- & --- & 114964.84$\pm$80421.84 / 92322.37 \\
\cellcolor{probmethod} RRT* & 12/90 & 45.23$\pm$14.85 / 42.48 & 2.75$\pm$4.04 / 3.22 & 2902.68$\pm$1524.33 / 2941.75 \\
\cellcolor{probmethod} RRT-C & 20/90 & 38.90$\pm$22.59 / 29.81 & 2.00$\pm$12.08 / -1.50 & 1530.38$\pm$1710.82 / 621.43 \\
\cellcolor{abstractmethod} \gls{fr} & 90/90 & 18.60$\pm$5.14 / 17.80 & 2.06$\pm$5.35 / 2.46 & 8114.62$\pm$3711.30 / 7772.22 \\
\cellcolor{abstractmethod} 2-stage PRM+A* & 90/90 & 11.34$\pm$15.31 / 6.43 & -2.75$\pm$7.07 / -1.26 & 12728.63$\pm$10851.24 / 10146.66 \\
\cellcolor{abstractmethod} HOPHY & 90/90 & \textbf{0.26$\pm$1.42 / 0.00} & \textbf{-0.05$\pm$0.85 / 0.00} & \textbf{661.52$\pm$231.07 / 629.80} \\
\bottomrule
\end{tabular}%
\endgroup
}
\end{table}

\subsection{Replanning under Weather Change}
\label{sec:replan_benchmark}
Having established static-query performance, we address Q2: whether HOPHY retains its speed and path quality after terrain conditions change. We test both distributed cost changes, representing rain-induced loss of friction, and a localized feasibility change, representing a newly blocked region.

\textbf{Setup}:
Using the same 90 pairs per map, light and medium rain reduce surface friction by landcover-specific fractions: 0.08/0.15 for forest, 0.10/0.20 for shrub, 0.15/0.25 for herbaceous, 0.20/0.30 for wetland, 0.12/0.22 for cropland, 0.05/0.10 for barren terrain, 0.03/0.06 for developed terrain, and 0.00/0.00 for water or snow. These are controlled sensitivity perturbations, not calibrated meteorological estimates. The blocked case instead makes one local region non-traversable. Every method receives the same modified raster and traversal objective.

The abstraction uses a weather hyperedge to identify affected \glspl{gsn}, updates their maintained state, and refines the new abstract route in its raster corridor. We compare with A* recomputed on the modified raster, 2-stage PRM+A*, and CLEAR using success, cost/length delta from A*, update time, and query time.

\begin{figure}[ht]
  \centering
  \vspace{-8pt}
  \includegraphics[width=0.8\linewidth]{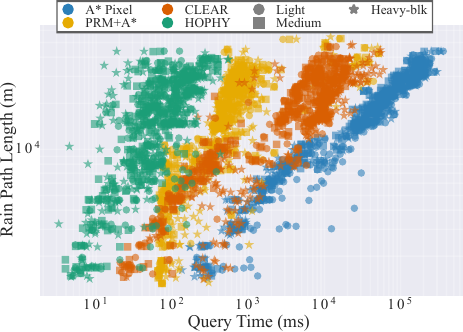}
\caption{Speed--quality trade-off for replanning. HOPHY path length stays close to the oracle (pixel A*) reference across rain and blockage effects.}
\vspace{-10pt}
  \label{fig:rain_speed_quality_tradeoff}
\end{figure}

Together,~\Cref{tab:replanning_efficiency} and~\Cref{fig:rain_speed_quality_tradeoff} show small path-quality deviations and sub-second mean replanning under the evaluated weather and blockage changes. Our abstraction completes all queries with zero median cost and length deviation. Even on Rainier, mean update-plus-query time remains below 400\,ms, compared with 89--102\,s per oracle query and 15.7--38.9\,s for CLEAR updates alone.

\begin{table}[t]
\centering
\tiny
\setlength{\tabcolsep}{1.4pt}
\renewcommand{\arraystretch}{0.82}
\caption{\rev{Replanning under Light Rain, Medium Rain, and Blocked conditions over 90 start--goal pairs per map/condition. Metrics are mean $\pm$ standard deviation / median; cost and length deltas are relative to saved pixel-A* paths. HOPHY update time measures context-state modification and cache invalidation, excluding hierarchy construction and static-descriptor preparation. Query time includes search, required cost evaluations, and raster refinement.}}
\label{tab:replanning_efficiency}
\resizebox{\columnwidth}{!}{%
\begingroup
\begin{tabular}{l l c c c c}
\toprule
\textbf{Condition} & \textbf{Method} &
\textbf{$\Delta$Cost (\%)} & \textbf{$\Delta$Len. (\%)} &
\textbf{Upd. (ms)} & \textbf{Query (ms)} \\
\midrule
\multicolumn{6}{l}{\textbf{Wharton (W)}} \\
Light Rain & \cellcolor{oraclemethod} A* & -- & -- & -- & 4340.4$\pm$5000.4/2703.5 \\
 & \cellcolor{abstractmethod} \gls{fr} & 3.3$\pm$2.1/2.8 & 0.4$\pm$4.5/1.0 & 3425.2$\pm$0.0/3425.2 & 232.7$\pm$376.8/135.4 \\
 & \cellcolor{abstractmethod} 2-stage PRM+A* & \textbf{0.0$\pm$0.0/0.0} & \textbf{0.0$\pm$0.0/0.0} & -- & 230.4$\pm$398.0/118.5 \\
 & \cellcolor{abstractmethod} HOPHY & \textbf{0.0$\pm$0.0/0.0} & \textbf{0.0$\pm$0.0/0.0} & \textbf{0.35$\pm$0.03/0.35} & \textbf{27.6$\pm$56.3/17.4} \\
\cmidrule(lr){1-6}
Medium Rain & \cellcolor{oraclemethod} A* & -- & -- & -- & 2139.1$\pm$1853.2/1544.4 \\
 & \cellcolor{abstractmethod} \gls{fr} & 2.7$\pm$1.5/2.5 & 0.8$\pm$3.7/0.6 & 3002.4$\pm$0.0/3002.4 & 231.8$\pm$306.4/139.1 \\
 & \cellcolor{abstractmethod} 2-stage PRM+A* & \textbf{0.0$\pm$0.0/0.0} & \textbf{0.0$\pm$0.0/0.0} & -- & 173.8$\pm$209.4/125.6 \\
 & \cellcolor{abstractmethod} HOPHY & \textbf{0.0$\pm$0.0/0.0} & \textbf{0.0$\pm$0.0/0.0} & \textbf{0.34$\pm$0.02/0.34} & \textbf{29.3$\pm$59.4/19.0} \\
\cmidrule(lr){1-6}
Blocked & \cellcolor{oraclemethod} A* & -- & -- & -- & 1959.8$\pm$1792.8/1473.1 \\
 & \cellcolor{abstractmethod} \gls{fr} & 1.8$\pm$3.3/1.8 & -8.3$\pm$10.6/-4.8 & 7906.3$\pm$0.0/7906.3 & 552.0$\pm$372.8/460.1 \\
 & \cellcolor{abstractmethod} 2-stage PRM+A* & \textbf{0.0$\pm$0.0/0.0} & \textbf{0.0$\pm$0.0/0.0} & -- & 1513.0$\pm$1115.3/1424.1 \\
 & \cellcolor{abstractmethod} HOPHY & \textbf{0.0$\pm$0.0/0.0} & \textbf{0.0$\pm$0.0/0.0} & \textbf{3.03$\pm$2.36/2.32} & \textbf{32.9$\pm$76.9/16.1} \\
\midrule
\multicolumn{6}{l}{\textbf{Humphrey (H)}} \\
Light Rain & \cellcolor{oraclemethod} A* & -- & -- & -- & 87531.0$\pm$75550.1/60220.0 \\
 & \cellcolor{abstractmethod} \gls{fr} & 15.8$\pm$3.4/15.4 & 4.0$\pm$3.1/3.6 & 10701.5$\pm$0.0/10701.5 & 7104.0$\pm$4496.7/6113.5 \\
 & \cellcolor{abstractmethod} 2-stage PRM+A* & 3.6$\pm$3.7/2.5 & 3.5$\pm$4.9/1.4 & -- & 4184.9$\pm$5733.6/2020.4 \\
 & \cellcolor{abstractmethod} HOPHY & \textbf{0.0$\pm$0.0/0.0} & \textbf{0.0$\pm$0.0/0.0} & \textbf{9.39$\pm$0.43/9.27} & \textbf{219.7$\pm$92.8/213.2} \\
\cmidrule(lr){1-6}
Medium Rain & \cellcolor{oraclemethod} A* & -- & -- & -- & 65902.1$\pm$57230.8/45483.3 \\
 & \cellcolor{abstractmethod} \gls{fr} & 14.8$\pm$3.1/14.5 & 4.1$\pm$3.3/3.9 & 11328.1$\pm$0.0/11328.1 & 6643.1$\pm$4340.1/5950.8 \\
 & \cellcolor{abstractmethod} 2-stage PRM+A* & 3.8$\pm$4.0/2.7 & 4.1$\pm$5.3/2.6 & -- & 3912.7$\pm$3682.9/2284.2 \\
 & \cellcolor{abstractmethod} HOPHY & \textbf{0.0$\pm$0.0/0.0} & \textbf{0.0$\pm$0.0/0.0} & \textbf{9.21$\pm$0.05/9.21} & \textbf{216.4$\pm$91.7/195.3} \\
\cmidrule(lr){1-6}
Blocked & \cellcolor{oraclemethod} A* & -- & -- & -- & 74552.3$\pm$66132.7/53651.4 \\
 & \cellcolor{abstractmethod} \gls{fr} & 20.7$\pm$5.3/20.6 & 6.3$\pm$4.6/6.3 & 27218.2$\pm$0.0/27218.2 & 11631.2$\pm$5788.0/10492.6 \\
 & \cellcolor{abstractmethod} 2-stage PRM+A* & 3.4$\pm$3.7/2.2 & 3.4$\pm$5.3/1.3 & -- & 15614.2$\pm$9563.8/13588.2 \\
 & \cellcolor{abstractmethod} HOPHY & \textbf{0.04$\pm$0.39/0.0} & \textbf{-0.03$\pm$0.31/0.0} & \textbf{4.21$\pm$2.55/3.64} & \textbf{213.5$\pm$87.6/203.0} \\
\midrule
\multicolumn{6}{l}{\textbf{Rainier (R)}} \\
Light Rain & \cellcolor{oraclemethod} A* & -- & -- & -- & 101822.6$\pm$61482.4/83354.8 \\
 & \cellcolor{abstractmethod} \gls{fr} & 20.3$\pm$4.9/20.0 & 1.0$\pm$5.0/0.9 & 16125.4$\pm$0.0/16125.4 & 8533.2$\pm$4248.2/8255.5 \\
 & \cellcolor{abstractmethod} 2-stage PRM+A* & 10.7$\pm$11.1/6.5 & 4.6$\pm$7.6/1.5 & -- & 14801.4$\pm$40478.0/6792.7 \\
 & \cellcolor{abstractmethod} HOPHY & \textbf{0.29$\pm$1.55/0.0} & \textbf{-0.06$\pm$0.81/0.0} & \textbf{26.98$\pm$0.13/27.01} & \textbf{368.6$\pm$173.2/319.6} \\
\cmidrule(lr){1-6}
Medium Rain & \cellcolor{oraclemethod} A* & -- & -- & -- & 88972.9$\pm$56990.9/70464.0 \\
 & \cellcolor{abstractmethod} \gls{fr} & 19.3$\pm$4.8/19.4 & 2.3$\pm$5.7/2.1 & 15744.1$\pm$0.0/15744.1 & 8394.5$\pm$4233.8/8057.1 \\
 & \cellcolor{abstractmethod} 2-stage PRM+A* & 10.2$\pm$10.7/6.1 & 5.2$\pm$7.6/2.2 & -- & 14776.3$\pm$38713.0/6933.3 \\
 & \cellcolor{abstractmethod} HOPHY & \textbf{0.28$\pm$1.53/0.0} & \textbf{-0.04$\pm$0.72/0.0} & \textbf{27.25$\pm$0.49/27.05} & \textbf{367.2$\pm$169.2/318.4} \\
\cmidrule(lr){1-6}
Blocked & \cellcolor{oraclemethod} A* & -- & -- & -- & 98131.4$\pm$62250.4/78354.8 \\
 & \cellcolor{abstractmethod} \gls{fr} & 21.7$\pm$12.8/23.2 & 3.8$\pm$7.5/4.5 & 38894.3$\pm$0.0/38894.3 & 18516.4$\pm$14146.0/13445.6 \\
 & \cellcolor{abstractmethod} 2-stage PRM+A* & 9.4$\pm$10.4/5.6 & 4.5$\pm$7.2/1.8 & -- & 27157.6$\pm$54059.7/12815.6 \\
 & \cellcolor{abstractmethod} HOPHY & \textbf{0.26$\pm$1.42/0.0} & \textbf{-0.05$\pm$0.85/0.0} & \textbf{7.57$\pm$3.31/7.62} & \textbf{362.2$\pm$165.8/311.3} \\
\bottomrule
\end{tabular}%
\endgroup
}
\end{table}

\subsection{Mechanism Analysis}
\label{sec:ablation}

\begin{figure}[t]
\centering
\includegraphics[width=\linewidth]{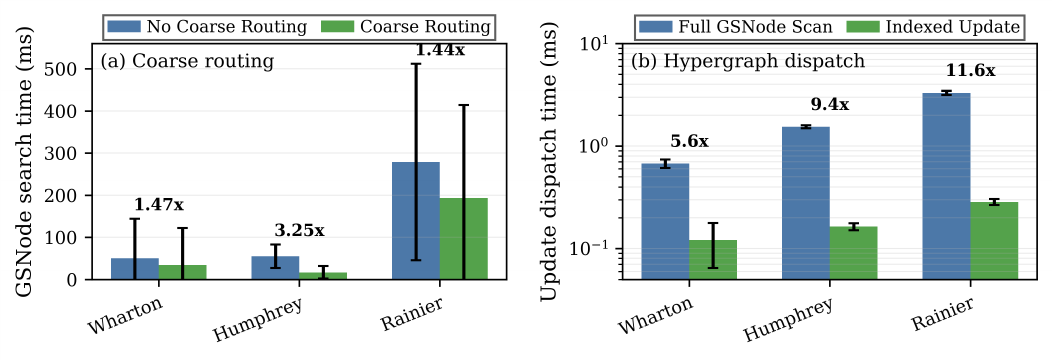}
\vspace{-11pt}
\caption{HOPHY ablation (mean$\pm$std, $n=90$). (a) GSNode search only. (b) Update dispatch only. Both variants apply identical node and incident-adjacency mutations.}
\vspace{-8pt}
\label{fig:ablation_study}
\end{figure}

Fig.~\ref{fig:ablation_study} reruns both mechanism ablations on HOPHY. Enabling \gls{cr} routing gives GSNode-search speedups of 1.47$\times$/3.25$\times$/1.44$\times$ on W/H/R. For the same 90 affected sets and state mutations, indexed dispatch avoids a full \gls{gsn} scan and is 5.6$\times$/9.4$\times$/11.6$\times$ faster.

\begin{figure}[t]
\centering
\includegraphics[width=0.8\linewidth]{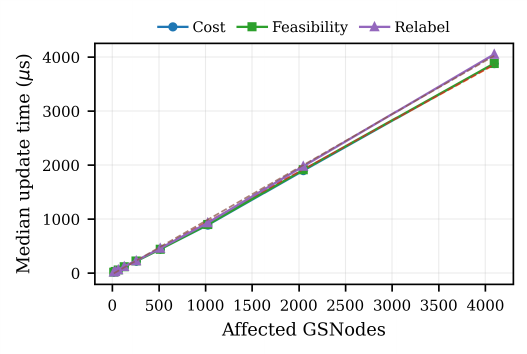}
\vspace{-7pt}
\caption{HOPHY hyperedge update time vs.\ affected GSNodes (three membership types). Cost, feasibility, and relabel operators update the matched nodes and incident edges.}
\vspace{-9pt}
\label{fig:typed_update_scaling}
\end{figure}

\noindent\textbf{Scaling with the affected set.}\label{sec:update_scaling}
We next vary the affected set from 16 to 4096 nodes on the direct 11{,}993-GSNode Wharton graph while fixing three membership types: terrain, agent, and weather. Their intersection identifies the changed subset. Every matched node and incident edge is then updated. Fig.~\ref{fig:typed_update_scaling} shows near-linear scaling, with exponents of 1.005--1.009 and $R^2\!\geq\!0.9995$ across cost, feasibility, and relabel operators.

\subsection{Multi-Agent Task Allocation}
\label{sec:coa_eval}
Q4 moves from individual queries to a downstream optimization that repeatedly requests path-cost estimates. On Wharton, we use HOPHY, \gls{fr}, 2-stage PRM+A*, and pixel A* to construct ordered task-to-task traversal-time and energy matrices for Het-MinMax-mTSP instances with 2 agents/20 tasks and 5 agents/50 tasks. An elitist genetic algorithm then minimizes makespan, $\min ( \max_{j} \{ T_{a_j} \} )$. Each setting uses 10 runs with 100 iterations, population 50, elite ratio 0.01, crossover probability 0.5, and mutation probability 0.1.

\begin{figure}[t]
\centering
\includegraphics[width=\linewidth]{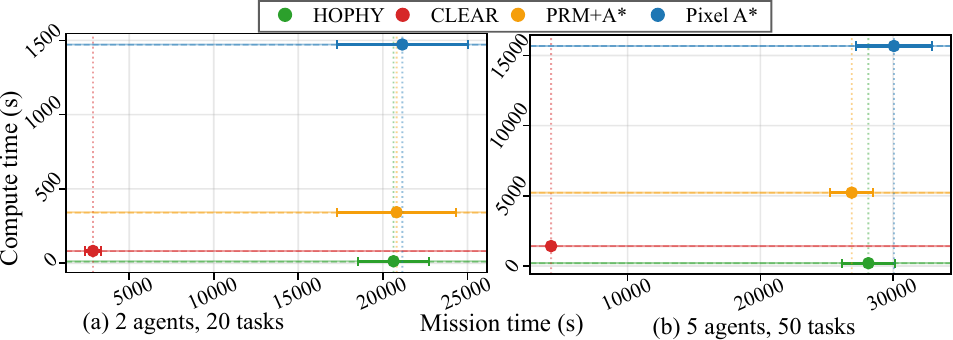}
\vspace{-10pt}
  \caption{Wharton MRTA results over 10 GA runs. Computation time is summed pairwise traversal-query time plus GA runtime, and mission completion time is the makespan estimated from each method's traversal-time matrix.}
  \label{fig:coa_demo}
\end{figure}

As shown in~\Cref{fig:coa_demo}, in the 5-agent/50-task case, HOPHY reduces computation from 15677 to 198\,s relative to pixel A* ($79\times$), from 5227 to 198\,s relative to 2-stage PRM+A* ($26\times$), and from 1416 to 198\,s relative to \gls{fr} ($7.2\times$), with mission makespan comparable to pixel A* and 2-stage PRM+A*. Makespans are reported in seconds and computed from each method's traversal-time matrix using the same allocation objective and GA settings. CLEAR's abstraction underestimates travel length relative to pixel A*, yielding optimistic travel-time and makespan estimates. HOPHY's pixel-level refinement better preserves route geometry and produces makespans close to the oracle. CLEAR's lower estimate therefore does not establish faster mission execution.

\subsection{Physical Jackal Feasibility Demonstration}
\label{sec:real_robot_anchor}

\begin{figure}[t]
\centering
\includegraphics[width=0.7\linewidth]{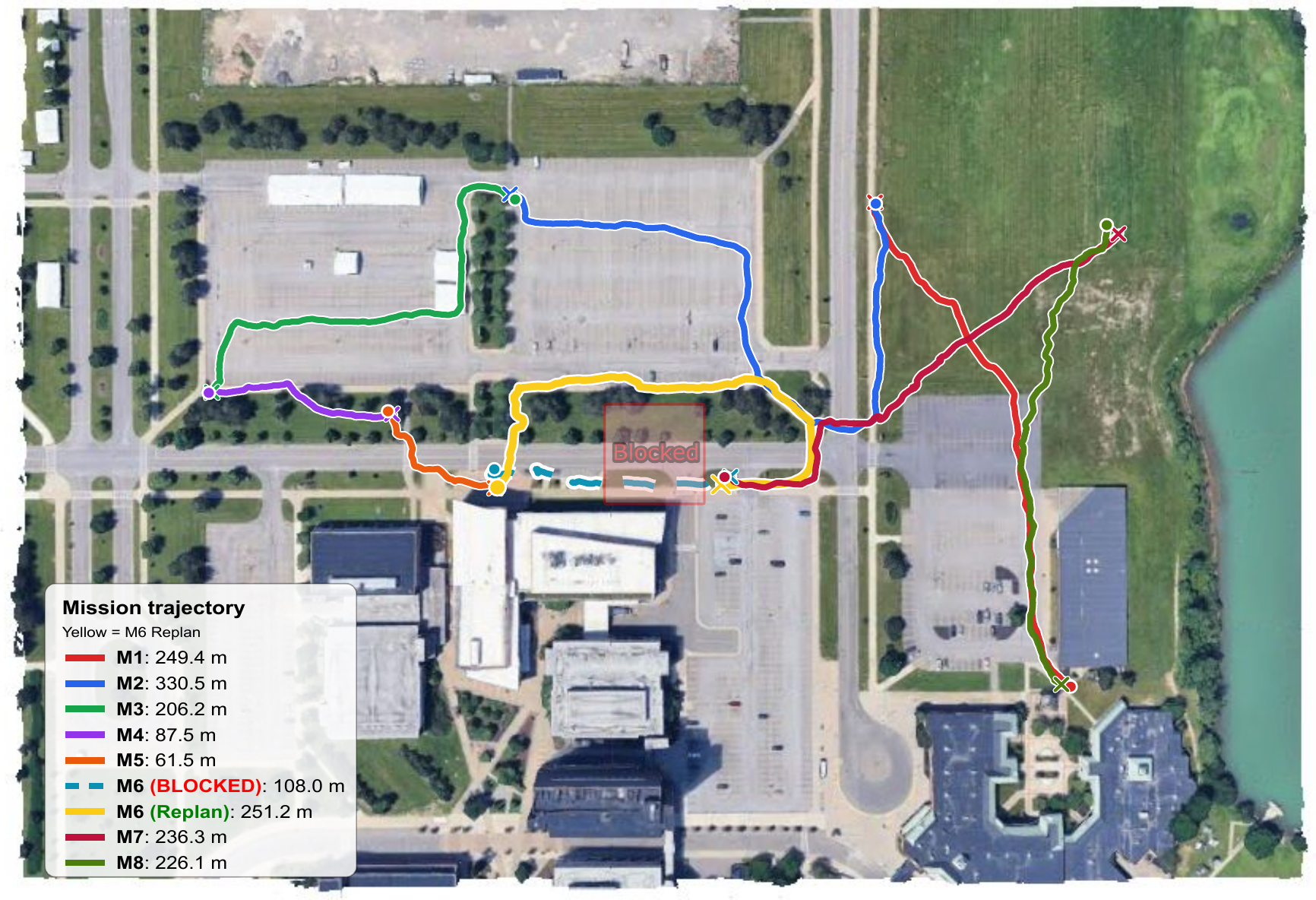}
\vspace{-6pt}
\caption{\rev{Jackal mission routes and blockage-triggered replanning at the mixed-terrain outdoor site. Dashed cyan: blocked M6 plan. Yellow: 251.2\,m replacement. Circles and crosses mark route endpoints.}}
\vspace{-8pt}
\label{fig:real_robot_anchor}
\end{figure}

Finally, Q5 tests whether paths generated using the proposed representation can be executed beyond the map benchmarks. We used the abstraction to generate a complete mission for a Clearpath Jackal spanning eight task locations across concrete, road, grass, and sloped muddy terrain. Mission generation required 2.65\,s, and the robot executed the resulting paths using Nav2. The aggregate feasibility outcomes are reported in~\Cref{tab:real_robot_mission}. \Cref{fig:real_robot_anchor} also illustrates replanning after we simulated a construction site that blocked the planned M6 route. Updating the representation required 67.5\,ms, and querying the replacement path required 137.1\,ms.

Following prior outdoor-robot field reporting~\cite{weerakoon_terp_2022}, an RTK-GNSS receiver with NTRIP corrections provided GPS waypoint localization with 5--15\,cm reported horizontal error. Nav2's Regulated Pure Pursuit controller followed the waypoints at 1.5\,m/s desired speed, while an Ouster LiDAR supported local obstacle detection.

\begin{table}[t]
\centering
\scriptsize
\setlength{\tabcolsep}{3.5pt}
\renewcommand{\arraystretch}{0.95}
\caption{Physical Jackal feasibility summary. The eight-segment mission and the replanned replacement route were each executed once, results are not repeated-trial statistics.}
\label{tab:real_robot_mission}
\begin{tabular}{l c r r}
\toprule
\rev{\textbf{Demonstration}} & \rev{\textbf{Outcome}} &
\rev{\textbf{Distance (m)}} & \rev{\textbf{Execution (s)}} \\
\midrule
\rev{Eight-segment mission} & \rev{8/8 completed} & \rev{1505.4} & \rev{994.5} \\
\rev{Blocked-route replan} & \rev{Replacement executed} & \rev{251.2} & \rev{164.4} \\
\bottomrule
\end{tabular}
\vspace{-4pt}
\end{table}
Across the paths, median map-derived slope ranges from 6.16$^\circ$ to 20.47$^\circ$, acceleration-based roughness RMS from 3.44 to 6.56\,m/s$^2$, and gyro RMS from 0.32 to 0.81\,rad/s. These ranges show that the robot successfully executed paths generated by the abstraction across terrain with substantially different steepness and motion-induced disturbance levels.

\section{Discussion \& Limitations}
\label{sec:discussion}

Across the evaluated queries, HOPHY exhibits small path-cost deviations from pixel A*. The cost model and hierarchy parameters are calibrated for off-road UGV planning. Transfer to other domains or unseen surface types warrants separate evaluation.

\textbf{When to Use}:
HOPHY is most beneficial when a terrain map is reused for repeated path queries, replanning, or mission optimization. Initialization takes approximately the same computation time as 28, 20, and 31 mean HOPHY queries on Wharton, Humphrey, and Rainier, respectively.

\textbf{Future Work}:
The maintained hypergraph backend opens several broader research directions. First, the representation is domain-agnostic in principle: applying it to urban, aerial, or multi-modal settings would require domain-appropriate segmentation and cost models, offering a path toward a unified planning backend across environments. Second, tighter integration with learning offers opportunities to replace fixed hierarchy parameters with data-driven construction and to update geometry online from onboard sensing, moving toward a self-maintaining terrain model. Third, scaling the typed update mechanism to distributed multi-agent teams—where agents hold partial views of a shared hypergraph—connects this work to decentralized planning and communication-aware mission optimization. Code, benchmark scripts, and trained model weights will be released with the paper.

\section{Conclusion}
\label{sec:conclusion}

We presented HOPHY, a reusable hierarchical terrain representation for off-road path and mission planning. GSNodes and Coarse Regions restrict route search, while typed hyperedge intersections identify terrain regions and incident edges affected by context changes. Across the evaluated maps, our representation supports fast repeated queries and replanning without rebuilding the hierarchy, with small path-cost deviations from pixel A*. MRTA experiments demonstrate reduced planning computation, and the Jackal deployment establishes execution feasibility. These results support persistent terrain representations as a planning backend for repeated mission-level queries.


\bibliographystyle{IEEEtran}
\bibliography{hhz_bib}

\end{document}